\documentclass{article}
\usepackage{spconf,amsmath,graphicx}

\usepackage{caption}
\usepackage{wrapfig}
\usepackage{setspace}
\usepackage{cleveref}

\usepackage{multirow}
\usepackage{xcolor}

\makeatletter
\renewcommand{\section}{\@startsection
  {section}%
  {1}%
  {}%
  {-0.5\baselineskip}%
  {0.2\baselineskip}%
  {}}%

\renewcommand{\subsection}{\@startsection
  {subsection}%
  {2}%
  {}%
  {-0.1\baselineskip}%
  {0.1\baselineskip}%
  {}}%

\renewcommand{\subsubsection}{\@startsection
  {subsubsection}%
  {3}%
  {}%
  {-0.2\baselineskip}%
  {0.2\baselineskip}%
  {}}%

\g@addto@macro\normalsize{%
  \setlength\abovedisplayskip{5pt plus 2pt minus 2pt}
  \setlength\belowdisplayskip{5pt plus 2pt minus 2pt}
  \setlength\abovedisplayshortskip{4pt plus 2pt minus 2pt}
  \setlength\belowdisplayshortskip{4pt plus 2pt minus 2pt}
}

\makeatother

\Crefname{equation}{Eq.}{Eqs.}
\Crefname{figure}{Fig.}{Figs.}
\Crefname{tabular}{Tab.}{Tabs.}

\DeclareMathOperator*{\argmax}{arg\,max\hspace{1mm}}

\newcommand\numberthis{\addtocounter{equation}{1}\tag{\theequation}}

\newcommand{\blank}{{\left<\operatorname{b}\right>}}

\title{Phoneme based Neural Transducer for Large Vocabulary Speech Recognition}
\name{Wei Zhou$^{1,2}$, Simon Berger$^{1}$, Ralf Schl\"uter$^{1,2}$, Hermann Ney$^{1,2}$}
\address{
$^1$Human Language Technology and Pattern Recognition, Computer Science Department,\\
  RWTH Aachen University, 52074 Aachen, Germany \\
$^2$AppTek GmbH, 52062 Aachen, Germany}

\begin{document}
%
\maketitle
\begin{abstract}
To join the advantages of classical and end-to-end approaches for speech recognition, we present a simple, novel and competitive approach for phoneme-based neural transducer modeling. Different alignment label topologies are compared and word-end-based phoneme label augmentation is proposed to improve performance. Utilizing the local dependency of phonemes, we adopt a simplified neural network structure and a straightforward integration with the external word-level language model to preserve the consistency of seq-to-seq modeling. We also present a simple, stable and efficient training procedure using frame-wise cross-entropy loss. A phonetic context size of one is shown to be sufficient for the best performance. A simplified scheduled sampling approach is applied for further improvement and different decoding approaches are briefly compared. The overall performance of our best model is comparable to state-of-the-art (SOTA) results for the TED-LIUM Release 2 and Switchboard corpora.
\end{abstract}
\vspace{-1mm}
\vspace{-1mm}
\section{Introduction \& Related Work}
\vspace{-1mm}
On reasonably sized automatic speech recognition (ASR) tasks like the TED-LIUM Release 2 (TLv2) \cite{tedlium2} and Librispeech \cite{libsp}, the classical hybrid hidden Markov model (HMM) \cite{bourlard1993hybridhmm} approach still shows superior performance \cite{zhou2020icassp, luescher2019librispeech}. Its composition of individual acoustic model (AM), lexicon and language model (LM) gives strong flexibility at the cost of complexity. Additionally, as phonetic units are usually used for the AM, the hybrid HMM also shows a good scalability to low-resource tasks. However, the formulation with conditional independence assumption and approximated prior as well as the clustered acoustic modeling units \cite{young93cart} lead to inconsistency of modeling.

Recently, the end-to-end approach, which enables the direct mapping of acoustic feature sequences to sub-word or word sequences, has shown competitive performance for ASR \cite{zoltan2020swb, zoph2019specaugment}. Common end-to-end models include connectionist temporal classification (CTC) \cite{graves2016ctc}, recurrent neural network transducer (RNN-T) \cite{graves2012sequence}, attention-based encoder-decoder models \cite{bahdanau2016end, chan2016listen} and possible variants thereof. The integration of all components into one powerful neural network (NN) for joint optimization leads to a great simplicity at the cost of less flexibility. Additionally, the straightforward seq-to-seq modeling results in more consistent training and inference. However, good performance usually requires large amount of training data and/or data augmentation \cite{zoph2019specaugment} as well as much longer training.

Attempts to join the advantages of both approaches have been arising. In \cite{wang2020phonAttention}, phoneme-based sub-word units are applied to attention models. Together with an external lexicon and both sub-word and word-level LMs, the overall system achieved SOTA results on the Switchboard task. More relevant to this work is the hybrid autoregressive transducer (HAT) \cite{variani2020hat}, which can be regarded as one variant of RNN-T. Using phonemes as label units, HAT formulates the problem into a pseudo generative modeling by exploring the model's internal LM and obtains large improvement over the baseline RNN-T. Additionally, limited phonetic context is shown to retain the performance of full context modeling, which is also verified by sub-word-based RNN-T \cite{ghodsi2020statelessrnnt}.

Following this motivation, we investigate phoneme-based neural transducer models in this work. We compare different alignment label topologies for transducer modeling and propose phoneme label augmentation to improve performance. Unlike HAT, no internal LM is applied in our approach as we claim that by modeling only the local dependency of phonemes (co-articulation), the negative effect of model's internal LM can be largely suppressed or avoided. This allows a straightforward integration with the external word-level LM, such as LM shallow fusion \cite{gulcehre2015shallowFusion}, and a simplified NN structure without the necessity of a separate blank distribution. Similar as in \cite{zeyer2020transducer}, we explore a simple, stable and efficient training procedure using frame-wise cross-entropy (CE) loss and apply a simplified scheduled sampling \cite{bengio2015scheduledSampling} approach to further improve performance. Different phonetic context sizes and decoding approaches are also investigated. Experiments on the TLv2 and 300h-Switchboard (SWBD) \cite{swb} corpora show that our best results are close to SOTA performance.

\vspace{-1mm}
\section{Phoneme-based Neural Transducer}
\vspace{-1mm}
\subsection{Model definition \& label topology}
Let $x_1^{T'}$ and $a_1^S$ denote the input feature sequence and output phoneme label sequence, respectively. And let $h_1^T = f^{\text{enc}}(x_1^{T'})$ denote the encoder output, which transforms the input into a sequence of high-level representations. In general, $T \le T'$ due to optional sub-sampling in the encoder and $T > S$. Let $y_1^U$ denote the alignment sequence between $h_1^T$ and $a_1^S$, where $U=T$ by introducing strict monotonicity (time-synchronous) as in \cite{sak2017rna, tripathi2019monoRNNT}. Each $y_1^U$ can be uniquely mapped to $a_1^S$ with an additional transition sequence $s_1^U$, where $y_u$ is mapped to $a_{s_u}$. Note that $s_u \in \{s_{u-1}, s_{u-1}+1\}$ and $0 \le s_u \le S$ at each alignment step $u$, where $s_u=0$ stands for no label yet. This also allows label repetition in $a_1^S$. The output label sequence posterior can be obtained as:\\
\scalebox{0.85}{\parbox{1.17\linewidth}{%
\begin{align*}
&p(a_1^S \mid x_1^{T'}) = \sum_{(y, s)_1^U:a_1^S} p(y_1^U, s_1^U \mid h_1^T) \numberthis \label{eq:seqPosterior}
\end{align*}}}\\
Here we compare two different alignment label topologies for neural transducer modeling, which further defines \Cref{eq:seqPosterior} into two different probabilistic modeling approaches.

\subsubsection{RNA topology}
\vspace{-1mm}
The first topology is the same as recurrent neural aligner (RNA) \cite{sak2017rna} or monotonic RNN-T \cite{tripathi2019monoRNNT}, where each $a_s$ appears only once in $y_1^U$ and the rest of $y_1^U$ is filled with the additional blank label $\blank$. Following \cite{zeyer2020transducer}, we call this the RNA topology. 
In this case, $y_1^U$ can fully define $s_1^U$, as $y_u=\blank$ represents $s_u = s_{u-1}$ and $y_u \neq \blank$ represents $s_u = s_{u-1}+1$. Thus, \Cref{eq:seqPosterior} can be directly simplified as:\\
\scalebox{0.85}{\parbox{1.17\linewidth}{%
\begin{align*}
p(a_1^S \mid x_1^{T'}) &= \sum_{(y, s)_1^U:a_1^S} \prod_{u=1}^U p(y_u \mid y_1^{u-1}, h_1^T) \\
&= \sum_{(y, s)_1^U:a_1^S} \prod_{u=1}^U p_{\theta}(y_u \mid a_{s_{u-1}-k+1}^{s_{u-1}}, h_1^T) \numberthis \label{eq:rna}
\end{align*}}}
where $p_{\theta}$ is the underlying parameterized decoder, which estimates a probability distribution over the full label vocabulary including $\blank$ based on the given label context and encoder output. Here we additionally introduce $k$ to define the context size. With $k=s_{u-1}$, \Cref{eq:rna} leads to the standard definition of RNN-T with full context and strict monotonicity. Theoretically, we can also introduce $u_s$ to denote positions in $y_1^U$ where $a_s$ occurs and reformulate \Cref{eq:rna} into a segmental modeling, which however, is not investigated in this work.

\subsubsection{HMM topology}
\vspace{-1mm}
The second one is the classical HMM topology which is widely used for alignment in the hybrid HMM approach. Instead of using $\blank$, each $a_s$ can loop for multiple steps in $y_1^U$, where additional non-speech labels, e.g. silence, may also be introduced for $a_s$. We can then define \Cref{eq:seqPosterior} as:\\
\scalebox{0.85}{\parbox{1.17\linewidth}{%
\begin{align*}
\sum_{(y, s)_1^U:a_1^S} \prod_{u=1}^U p(s_u \mid y_1^{u-1}, s_1^{u-1}, h_1^T) \cdot p(y_u \mid y_1^{u-1}, s_1^{u}, h_1^T)
\end{align*}}}
where $p(s_u \mid y_1^{u-1}, s_1^{u-1}, h_1^T)$ is defined as:\\
\scalebox{0.85}{\parbox{1.17\linewidth}{%
\begin{align*}
\begin{cases}
q_{\theta}(y_u=y_{u-1} \mid a_{s_{u-1}-k}^{s_{u-1}-1}, h_1^T), &s_u = s_{u-1} \\
1 - q_{\theta}(y_u=y_{u-1} \mid a_{s_{u-1}-k}^{s_{u-1}-1}, h_1^T), &s_u = s_{u-1}+1 \numberthis \label{eq:nonloop}
\end{cases}
\end{align*}}}
and $p(y_u \mid y_1^{u-1}, s_1^u, h_1^T)$ is defined as:\\
\scalebox{0.85}{\parbox{1.17\linewidth}{%
\begin{align*}
\begin{cases}
\delta_{y_u, y_{u-1}}, &s_u = s_{u-1} \\
q_{\theta}(y_u \mid a_{s_u-k}^{s_u-1}, h_1^T), &s_u = s_{u-1}+1
\end{cases}
\end{align*}}}
Similarly, $q_{\theta}$ is the underlying parameterized decoder. Note that at forward transitions, this definition first computes a non-loop probability as in \Cref{eq:nonloop} using the previous context, which is necessary for proper normalization, and then computes the next label probability with the updated context.

\subsection{Decision and decoding}
\vspace{-1mm}
Together with an external word-level LM and lexicon, the final best word sequence can be decided as:\\
\scalebox{0.85}{\parbox{1.17\linewidth}{%
\begin{align*}
\hspace{-2mm} x_1^{T'} \rightarrow \tilde{w}_1^N &= \argmax_{w_1^N} p^{\lambda}(w_1^N) \sum_{a_1^S:w_1^N} p(a_1^S \mid x_1^{T'}) \\
&= \argmax_{w_1^N} p^{\lambda}(w_1^N) \sum_{{(y,s)}_1^U:a_1^S:w_1^N} p(y_1^U, s_1^U \mid h_1^T) \numberthis \label{eq:fullsum}\\
&\approx \argmax_{w_1^N} p^{\lambda}(w_1^N) \max_{{(y,s)}_1^U:a_1^S:w_1^N} p(y_1^U, s_1^U \mid h_1^T) \numberthis \label{eq:viterbi}
\end{align*}}}
where $\lambda$ is the LM scale. Similar as in phoneme-based attention model \cite{tara2018icassp}, this decision rule simply adopts a log-linear model combination to maximumly keep the consistency of seq-to-seq modeling, which we extend into the framework of phoneme-based transducer model. \Cref{eq:fullsum} and \Cref{eq:viterbi} correspond to full-sum and Viterbi decoding, respectively.

We apply lexical prefix tree search with score-based pruning and optional LM look-ahead \cite{ortmanns1996lmla}. For the local dependency of phonemes, we set $k=1$ by default for our phoneme transducer models. This allows us to compute the scores of all possible label context at each step $u$ in a single batch forwarding, which are then cached for efficient reuse in decoding. Hypotheses are recombined, either summation or maximization, based on model and decoding settings.

\subsection{Label augmentation}
\label{sec:label}
\vspace{-1mm}
Word boundary information has been adopted to improve ASR performance for both hybrid HMM approach \cite{liao2010wbdt, le2019chenones} and phoneme-based end-to-end approach \cite{tara2018icassp, irie2019phoneme}. The latter inserted a separate end-of-word (EOW) label to the phoneme inventory, which however, does not correspond to any acoustic realization. We propose to augment the phoneme inventory with EOW discrimination by identifying each phoneme appearing at word end to be a different class than that appearing within the word. This label augmentation effectively increases the size of the label vocabulary by a factor of two, which is still very small and simple for phonemes. Besides, we also investigate the effect of applying the same for start-of-word (SOW) in addition to EOW. This increases the vocabulary size by a factor of four due to phonemes appearing at both SOW and EOW, i.e. single-phoneme pronunciation.
 
\subsection{Simplified NN architecture \& training}
\vspace{-1mm}
We use a simplified NN architecture derived from the RNN-T network structure \cite{graves2012sequence}. The encoder contains 6 bidirectional long short-term memory \cite{hochreiter1997lstm} (BLSTM) layers with 512 units for each direction. We apply sub-sampling by a factor of 2 via max-pooling after the third stack of BLSTM layers. A small context size $k$ leads to a feed-forward neural network (FFNN)-based instead of recurrent neural network (RNN)-based decoder. We use label embedding of size 128 and 2 linear layers of size 1024 and tanh activation. The encoder and FFNN outputs are simply summed up and fed to a final softmax layer to predict the posterior distribution over the full label vocabulary. This simplified NN structure is used throughout this work for both label topologies.

Standard training of transducer models requires full-sum over all possible alignments on the whole sequence, which can be both time and memory consuming. Similar as in \cite{zeyer2020transducer}, we apply Viterbi approximation to train our transducer model using frame-wise CE loss w.r.t. $p(y_1^U, s_1^U \mid h_1^T)$ and a fixed external alignment. For the HMM topology, this also includes the non-loop probability in \Cref{eq:nonloop} at forward transitions. For the alignment generation, as a simple pre-stage of training, we mainly consider hybrid HMM and CTC models, both of which are easy and fast to obtain. Such simplification also allows us to apply additional techniques to further speed up training and improve performance (\Cref{sec:exp}).

\vspace{-1mm}
\section{Experiments}
\label{sec:exp}
\vspace{-1.5mm}
\subsection{Setup}
\vspace{-1mm}
Experimental evaluation is done on the TLv2 \cite{tedlium2} and SWBD \cite{swb} corpora with the official lexicons (TLv2: 39 phonemes and 152k words; SWBD: 45 phonemes and 30k words). An additional silence label is used for the HMM topology, while a blank label for the RNA topology. For SWBD, the Hub5'00 and Hub5'01 datasets are used as dev and test set, respectively. All the LMs used are the same as in \cite{zhou2020icassp} for TLv2 and \cite{irie2019asru} (sentence-wise) for SWBD. By default, word error rate (WER) results are obtained with full-sum decoding and a 4-gram LM.

We extract gammatone features from 25ms windows with 10ms shift (TLv2: 50-dim; SWBD: 40-dim). By default, we use the hybrid HMM alignment and treat the last frame of each phoneme segment as $u_s$ for the RNA topology. To speed up training, sequences are decomposed into chunks (TLv2: 256 frames; SWBD: 128 frames) with 50\% overlap and a mini-batch of 128 chunks is used. An all-0 embedding is used for the initial computation. SpecAugment \cite{zoph2019specaugment} is applied as done in \cite{zhou2020icassp}.  We firstly pretrain the  encoder with frame-wise CE loss \cite{hu2020pretrain} for about 5 full epochs and then keep this encoder loss in further training with a focal loss factor 1.0 \cite{lin2017focalloss}. We use the Nadam optimizer \cite{nadam} with initial learning rate (LR) 0.001, which is kept constant for about 6 full epochs after pre-training. Then the Newbob LR scheduling \cite{zeyer2017newbob} with a decay factor of 0.9 and a minimum LR (TLv2: $2e^{-5}$; SWBD: $1e^{-5}$) are applied. All models converge well within about 50 full epochs in total. Additionally for the output CE loss of the RNA topology, we apply 0.2 label smoothing \cite{szegedy2016labelsmooth}, and boost the loss at positions $u_s$ by a factor of 5 to balance the large number of blank frames in the alignment.

\subsection{Label unit \& topology}
\vspace{-1mm}
We firstly compare the RNA topology and the HMM topology for neural transducer modeling. For both topologies, we evaluate the original, EOW-augmented and SOW+EOW-augmented phoneme label units (\Cref{sec:label}). Our label augmentation is applied only to speech phonemes. \Cref{tab:label} shows the WER results for both TLv2 and SWBD. For all three types of labels, the RNA topology shows consistently much better performance than the HMM topology, which appears to be more suitable for transducer modeling. In all cases, the EOW-augmented phoneme labels clearly improve over the original phoneme labels. Applying SOW in addition to EOW brings no further improvement but small degradation. This is intuitively clear as the SOW information is redundant for the model given the predecessor label with EOW information. Also the additional separation of single phoneme pronunciations might result in data sparsity problem. We use the EOW-augmented phoneme labels and the RNA topology for all further investigations.

\subsection{Alignment}
\vspace{-1mm}
For CE training using the hybrid HMM alignment, we can theoretically select any frame within each phoneme segment to be $u_s$ and treat the rest including silence to be $\blank$. Here we compare three different within-segment positions for $u_s$, namely, the first frame (segBeg), the middle frame (segMid) and the last frame (segEnd). Additionally, we also evaluate the alignment generated by phoneme-based CTC models. In this case, the alignment is really peaky as more than 95\% of the phoneme labels only consume one or two frames. And more than 30\% of them do not overlap with the corresponding segment in the hybrid HMM alignment. We simply choose segEnd for $u_s$. The WER results are shown in \Cref{tab:alignment}. For both corpora, our default setup performs the best. The difference among different cases is also not large, which can be further closed by more careful tuning. This suggests that our training procedure is rather stable w.r.t. different alignments of different properties.

\begin{table}[t!]
\caption{\it WER of different label units and topologies.}
\vspace{-1.5mm}
\small
\setlength{\tabcolsep}{0.2em}
\begin{center}\label{tab:label}
\begin{tabular}{|c|c|c|c|c|}
\hline
\multirow{2}{*}{Label} & \multicolumn{2}{c|}{TLv2-dev} & \multicolumn{2}{c|}{Hub5'00} \\ \cline{2-5}
 & RNA & HMM & RNA & HMM \\ \hline
phoneme & 7.6 & 9.3 & 14.0 & 15.4 \\ \hline 
+ EOW   & \textbf{6.9} & 8.8 & \textbf{13.4} & 14.5 \\ \hline
\quad + SOW & 7.3 & 9.0 & 13.5 & 14.8 \\ \hline
\end{tabular}
\end{center}
\vspace{-1mm}
\end{table}

\begin{table}[t!]
\caption{\it WER of different alignments and $u_s$ positions.}
\vspace{-1.5mm}
\small
\setlength{\tabcolsep}{0.2em}
\begin{center}\label{tab:alignment}
\begin{tabular}{|c|c|c|c|}
\hline
Alignment & $u_s$ & TLv2-dev & Hub5'00 \\ \hline
\multirow{3}{*}{\shortstack[c]{hybrid\\HMM}} & segBeg & 7.2 & 13.7 \\ \cline{2-4}
 & segMid & 7.4 & 13.8 \\ \cline{2-4}
 & \multirow{2}{*}{segEnd} & \textbf{6.9} & \textbf{13.4} \\ \cline{1-1} \cline{3-4}
CTC & & 7.2 & \textbf{13.4} \\ \hline 
\end{tabular}
\end{center}
\vspace{-5.5mm}
\end{table}


\subsection{Context}
\vspace{-1mm}
We also investigate the necessity of larger phoneme context size by varying $k \in \{1, 2, \infty\}$. For $k=\infty$, we replace the linear layers in the decoder with LSTM layers and no chunking is applied in training. The results are shown in \Cref{tab:context}. For CE loss with chunk-wise training, increased context size is not helpful, possibly due to the context loss at beginning of each chunk. Without chunking, increasing $k$ from 1 to 2 is beneficial for both corpora. However, the results are consistently worse than those with chunking. Additionally, performance degradation is observed when further increasing $k$ to $\infty$. This can result from the increasing effect of model's internal LM with larger context as pointed out in \cite{variani2020hat}, which is avoided with small context size. We also include results of standard RNN-T with full-sum (FS) training under similar amount of epochs. We initialize the encoder with a converged CTC model, but the training is  still very sensitive to different hyper-parameter settings. Our default setup is both better and much more efficient than the FS training.

\subsection{Ablation study and scheduled sampling}
\vspace{-1mm}
We perform ablation study on several major techniques applied in training, which include SpecAugment, chunking, encoder loss, label smoothing and the proposed loss boost at positions $u_s$ ($\text{lossBoost}_{u_s}$). As shown in \Cref{tab:ablation}, all of them are very helpful for our CE training, especially the $\text{lossBoost}_{u_s}$. 

Additionally, we also investigate the effect of scheduled sampling \cite{bengio2015scheduledSampling}. Without recurrence in the decoder NN, we simplify the approach as the following. For each chunk, we firstly feed the ground truth label context (alignment) into the network to produce posterior distribution for each frame. Then we randomly select 50\% of the frames to apply sampling based on the distribution. These sampled labels and the other 50\% ground truth labels are then jointly re-fed into the network for computing the output CE loss w.r.t. the alignment. We apply this simplified sampling approach to the converged models and reset the LR to further train them for maximum 50 epochs. The results are also shown in \Cref{tab:ablation}. We get no improvement for TLv2 but additional 4\% relative improvement for SWBD, which suggests that this approach can be more beneficial for more difficult tasks.

\begin{table}[t!]
\parbox{.55\linewidth}{
\caption{\it WER of different context and training; and training time in min/epoch [m/ep] on a single GTX 1080 Ti GPU.}
\vspace{-2mm}
\small
\setlength{\tabcolsep}{0.1em}
\begin{center}\label{tab:context}
\begin{tabular}{|c|c|c|c|c|c|c|}
\hline
\multirow{2}{*}{Loss} & \multirow{2}{*}{Chunk} & \multirow{2}{*}{$k$} & \multicolumn{2}{c|}{TLv2-dev} & \multicolumn{2}{c|}{Hub5'00} \\ \cline{4-7}
 &  &                       &  WER & m/ep & WER & m/ep \\ \hline
\multirow{5}{*}{CE} & \multirow{2}{*}{yes} & 1 & \textbf{6.9} & 93 & \textbf{13.4} & 132 \\ \cline{3-7}
& & 2 & 7.0 & \multirow{4}{*}{n.a.} & 13.6 & \multirow{4}{*}{n.a.} \\ \cline{2-4} \cline{6-6}
& \multirow{4}{*}{no} & 1 & 7.2 & & 14.1 & \\ \cline{3-4} \cline{6-6}
& & 2 & 7.0 & & 13.8 & \\ \cline{3-4} \cline{6-6}
& & \multirow{2}{*}{$\infty$} & 7.9 & & 15.3 & \\ \cline{1-1} \cline{4-7}
FS & & & 8.7 & 250 & 16.4 & 372\\ \hline
\end{tabular}
\end{center}
}
\hfill
\parbox{.4\linewidth}{
\caption{\it WER of ablation study and sampling.}
\vspace{-1mm}
\small
\setlength{\tabcolsep}{0.1em}
\begin{center}\label{tab:ablation}
\begin{tabular}{|l|c|c|}
\hline
\multirow{2}{*}{Training} & TLv2 & Hub \\ 
                          & dev & 5'00 \\ \hline
default       & 6.9 & 13.4 \\ \hline
- SpecAugment & 8.5 & 14.6 \\ \hline
- chunking    & 7.2 & 14.1 \\ \hline
- encoder loss & 7.3 & 14.0 \\ \hline
- label smooth & 8.0 & 14.2 \\ \hline
- $\text{lossBoost}_{u_s}$ & 9.9 & 16.4\\ \hline 
\hline
+ sampling & 6.9 & \textbf{12.9} \\ \hline
\end{tabular}
\end{center}
}
\vspace{-5.5mm}
\end{table}

\vspace{-3mm}
\begin{table}[h!]
\caption{\it WER of different decoding.}
\vspace{-1mm}
\small
\setlength{\tabcolsep}{0.2em}
\begin{center}\label{tab:decoding}
\begin{tabular}{|c|c|c|}
\hline
Decoding & TLv2-dev & Hub5'00 \\ \hline
Viterbi & 7.1 & 13.0 \\ \hline
full-sum & 6.9 & 12.9 \\ \hline
\end{tabular}
\end{center}
\vspace{-5.5mm}
\end{table}

\subsection{Decoding \& overall performance}
\vspace{-1mm}
We also check the effect of switching from full-sum to Viterbi decoding for our CE-trained model, which can enable additional word-end recombination to further simplify search. As shown in \Cref{tab:decoding}, there is minor but consistent degradation, which is also observed for hybrid HMM systems \cite{zhou2020fullsum}.

For our best model, we apply one-pass recognition with an LSTM LM to generate lattices that are further used for rescoring with a Transformer (Trafo) \cite{vaswani2017transformer} LM. The results are shown in \Cref{tab:tlv2} for TLv2 and \Cref{tab:swbd} for SWBD. We also include other results from the literature, which cover different modeling approaches using different labels. For TLv2, our best result is very close to the SOTA performance \cite{zhou2020icassp}, which applied additional speaker adaptive training and sequence discriminative training. For SWBD, our best result is still a little behind the SOTA performance \cite{zoltan2020swb}, which however, used much more epochs for training.

\begin{table}[t!]
\caption{\it Overall WER on TLv2 and results from literature.}
\vspace{-1.5mm}
\small
\setlength{\tabcolsep}{0.2em}
\begin{center}\label{tab:tlv2}
\begin{tabular}{|c|c|c|c|c|c|c|}
\hline
\multirow{2}{*}{Work} & \multicolumn{3}{c|}{Modeling} & \multirow{2}{*}{LM} & \multicolumn{2}{c|}{TLv2}\\ \cline{2-4} 
 & \#Epoch & Approach & Label &  & dev & test \\ \hline
\cite{karita2019asru} & 100 & Attention & sub-word & \multirow{2}{*}{RNN} & 9.3 & 8.1 \\ \cline{1-4} \cline{6-7}
\cite{han2018capio} & - & \multirow{3}{*}{\shortstack[c]{hybrid\\HMM}} & \multirow{3}{*}{triphone} & & 7.1 & 7.7 \\ \cline{1-2} \cline{5-7}
\multirow{2}{*}{\cite{zhou2020icassp}}  & \multirow{2}{*}{35} & & & LSTM & 5.6 & 6.0 \\ \cline{5-7}
                                      & & & & Trafo & 5.1 & 5.6 \\ \hline \hline
 \multirow{2}{*}{this} & \multirow{2}{*}{50} & \multirow{2}{*}{Transducer} & \multirow{2}{*}{phoneme} & LSTM & 5.9 & 6.3 \\ \cline{5-7}
            & & & & Trafo & 5.4 & 6.0 \\ \hline
\end{tabular}
\end{center}
\vspace{-1mm}
\end{table}

\begin{table}[t!]
\caption{\it Overall WER on SWBD and results from literature.}
\vspace{-1.5mm}
\small
\setlength{\tabcolsep}{0.2em}
\begin{center}\label{tab:swbd}
\begin{tabular}{|c|c|c|c|c|c|c|c|}
\hline
\multirow{2}{*}{Work} & \multicolumn{3}{c|}{Modeling} & \multirow{2}{*}{LM} & \multirow{2}{*}{\shortstack[c]{Hub\\5'00}} & \multirow{2}{*}{\shortstack[c]{Hub\\5'01}} \\ \cline{2-4}
                      & \#Epoch & Approach & Label & &  &  \\ \hline                   
\cite{zeyer2020transducer} & 50 & Transducer & sub-word & - & 13.5 & 13.3 \\ \hline
\cite{raissi2020interspeech} & 90 & hybrid HMM & phoneme-state & LSTM & 11.7 & - \\ \hline
\cite{zoph2019specaugment} & 760 & \multirow{2}{*}{Attention} & \multirow{2}{*}{sub-word} & RNN & 10.5 & - \\ \cline{1-2} \cline{5-7}
\cite{zoltan2020swb} & 250 & &  &  LSTM & \phantom{1}9.8 & 10.1 \\ \hline \hline
\multirow{2}{*}{this} & \multirow{2}{*}{100} & \multirow{2}{*}{Transducer} & \multirow{2}{*}{phoneme} & LSTM & 11.5 & 11.5 \\ \cline{5-7}
&  &  &  & Trafo & 11.2 & 11.2\\ \hline
\end{tabular}
\end{center}
\vspace{-5.5mm}
\end{table}

\vspace{-2mm}
\section{Conclusion}
\vspace{-1.5mm}
In this work, we presented a simple, novel and competitive approach for phoneme-based neural transducer modeling, which preserves advantages of both classical and end-to-end systems. By utilizing the local dependency of phonemes, we adopted a simplified NN structure and a straightforward integration with the external word-level LM to maintain the consistency of modeling. We also described a detailed training pipeline allowing a simple, stable and efficient training of transducer models using frame-wise CE loss. The RNA label topology is shown to be more suitable for transducer modeling than the HMM topology. The proposed EOW-augmented phoneme labels bring consistent improvement over the original phoneme set. A phonetic context size of one is shown to be sufficient for the best performance with chunk-wise training. The simplified sampling approach brings further improvement on the converged model for SWBD. We also briefly compared different decoding approaches. The overall performance of our best model is on par with SOTA results for TLv2 and a little behind SOTA performance for SWBD.

\vspace{-2mm}
\section{Acknowledgements}
\vspace{-1mm}
\begin{wrapfigure}[4]{l}{0.08\textwidth}
        \vspace{-4mm}
        \begin{center}
                \includegraphics[width=0.1\textwidth]{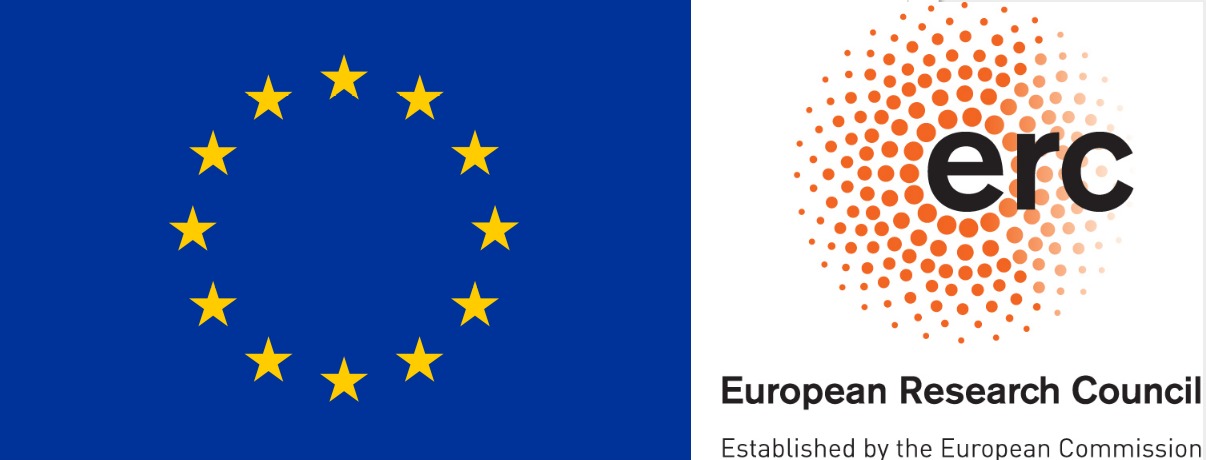} \\
        \end{center}
        \vspace{-4mm}
\end{wrapfigure}
\scriptsize
\setstretch{0.8}
This work has received funding from the European Research Council (ERC) under the European Union's Horizon 2020 research and innovation program (grant agreement No 694537, project ``SEQCLAS") and from a Google Focused Award. The work reflects only the authors' views and none of the funding parties is responsible for any use that may be made of the information it contains.

We thank Alexander Gerstenberger for the lattice rescoring experiments and Albert Zeyer for useful discussion.

\let\normalsize\small\normalsize
\let\OLDthebibliography\thebibliography
\renewcommand\thebibliography[1]{
        \OLDthebibliography{#1}
        \setlength{\parskip}{-0.3pt}
        \setlength{\itemsep}{1pt plus 0.07ex}
}

\bibliographystyle{IEEEbib}
\bibliography{refs}

\end{document}